\documentclass[runningheads]{llncs}

\usepackage[T1]{fontenc}

\usepackage{marvosym}
\usepackage[colorlinks,linkcolor=blue,anchorcolor=blue,citecolor=blue]{hyperref}

\usepackage{algorithmicx,algorithm}
\usepackage[noend]{algpseudocode}
\usepackage{bm}
\usepackage{amsmath}
\usepackage{multirow}
\usepackage{amssymb}     

\usepackage{graphicx}

\usepackage{color}

\begin{document}
\title{Partial Vessels Annotation-based Coronary Artery Segmentation with Self-training and Prototype Learning}

\titlerunning{Partial Vessels Annotation-based Coronary Artery Segmentation}
%
\author{Zheng Zhang\inst{1} \and Xiaolei Zhang\inst{2} \and Yaolei Qi\inst{1} \and Guanyu Yang\inst{1,3,4}(\Letter)}

\renewcommand{\thefootnote}{}
\footnotetext{Z. Zhang and X. Zhang\textemdash Contributed equally to this work.}
\authorrunning{Z. Zhang et al.}
%
\institute{LIST, Key Laboratory of Computer Network and Information Integration, Southeast University, Ministry of Education, Nanjing 210096, China
\email{yang.list@seu.edu.cn}\and
Dept. of Diagnostic Radiology, Jinling Hospital, Medical School of Nanjing University, Nanjing, China \and 
Jiangsu Provincial Joint International Research Laboratory of Medical
Information Processing, Southeast University, Nanjing 210096, China \and
Centre de Recherche en Information Biom´edicale Sino-Fran¸cais (CRIBs), Strasbourg, France
}
\maketitle              

\begin{abstract}
Coronary artery segmentation on coronary-computed tomography angiography (CCTA) images is crucial for clinical use. Due to the expertise-required and labor-intensive annotation process, there is a growing demand for the relevant label-efficient learning algorithms. To this end, we propose partial vessels annotation (PVA) based on the challenges of coronary artery segmentation and clinical diagnostic characteristics. Further, we propose a progressive weakly supervised learning framework to achieve accurate segmentation under PVA. First, our proposed framework learns the local features of vessels to propagate the knowledge to unlabeled regions. Subsequently, it learns the global structure by utilizing the propagated knowledge, and corrects the errors introduced in the propagation process. Finally, it leverages the similarity between feature embeddings and the feature prototype to enhance testing outputs. Experiments on clinical data reveals that our proposed framework outperforms the competing methods under PVA (24.29\% vessels), and achieves comparable performance in trunk continuity with the baseline model using full annotation (100\% vessels). 


\end{abstract} 

\section{Introduction}

 Coronary artery segmentation is crucial for clinical coronary artery disease diagnosis and treatment \cite{gharleghi2022towards}. Coronary-computed tomography angiography (CCTA), as a non-invasive technique, has been certified and recommended as established technology in the cardiological clinical arena \cite{serruys2021coronary}. Thus, automatic coronary artery segmentation on CCTA images has become increasingly sought after as a means to enhance diagnostic efficiency for clinicians. In recent years, the performance of deep learning-based methods have surpassed that of conventional machine learning approaches (e.g. region growing) in coronary artery segmentation \cite{gharleghi2022towards}. Nevertheless, most of these deep learning-based methods highly depend on accurately labeled datasets, which need labor-intensive annotations. Therefore, there is a growing demand for relevant label-efficient learning algorithms for automatic coronary artery segmentation on CCTA images.
 
Label-efficient learning algorithms have garnered considerable interest and research efforts in natural and medical image processing \cite{he2022learning,he2020dense,10048555}, while research on label-efficient coronary artery segmentation for CCTA images is slightly lagging behind. Although numerous label-efficient algorithms for coronary artery segmentation in X-ray angiograms have been proposed \cite{zhang2021ss,zhang2020weakly}, only a few researches focus on CCTA images. Qi et al. \cite{qi2021examinee} proposed an elabrately designed EE-Net to achieve commendable performance with limited labels. Zheng et al \cite{zheng2022ugan} transformed nnU-Net into semi-supervised segmentation field as the generator of Gan, having achieved satisfactory performance on CCTA images. Most of these researches use incomplete supervision, which labels a subset of data. However, other types of weak supervision (e.g. inexact supervision), which are widely used in natural image segmentation \cite{10048555}, are seldom applied to coronary artery segmentation on CCTA images.

\begin{figure*}
	\centering
	\includegraphics[width=\linewidth]{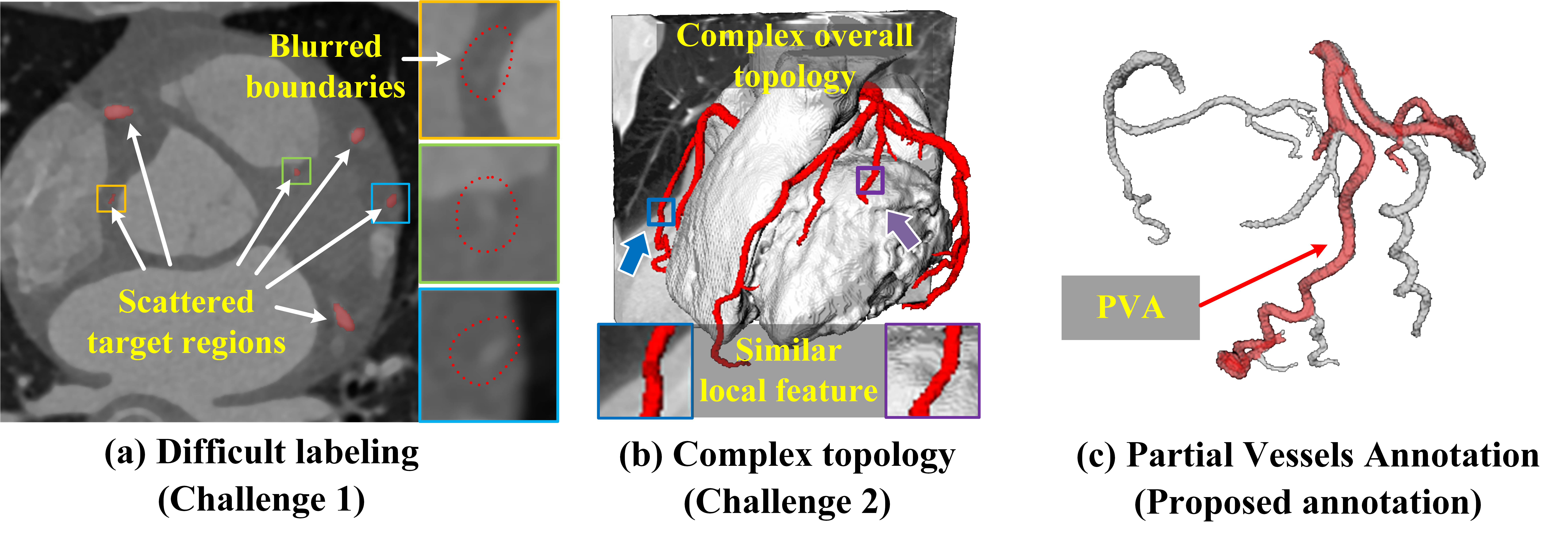}
	\caption{\textbf{Motivation}. (a) and (b) shows the two challenges of coronary artery segmentation, while (c) shows our proposed partial vessels annotation (PVA) according to the challenges. a) Coronary artery has blurred boundaries and scattered target regions. b) Coronary artery has complex overall topology but similar local feature. c) Partial vessels annotation (red) labels less regions than full annotation (overall).}
	\label{fig_diff}
\end{figure*}

Different types of supervision are utilized according to the specific tasks. The application of various types of weak supervision are inhibited in coronary artery segmentation on CCTA images by the following challenges. 1) Difficult labeling (Fig.~\ref{fig_diff}(a)). The target regions are scattered, while manual annotation is drawn slice by slice on the planes along the vessels. Also, boundaries of branches and peripheral vessels are blurred. These make the annotating process time-consuming and expertise-required. 2) Complex topology (Fig.~\ref{fig_diff}(b)). Coronary artery shows complex and slender structures, diameter of which ranges from 2 mm to 5 mm. The tree-like structure varies individually. Based on these challenges and the insight that vessels share local feature (Fig.~\ref{fig_diff}(b)), we propose partial vessels annotation and our framework as following.

Given the above, we propose partial vessels annotation (PVA) (Fig.~\ref{fig_diff}(c)) for CCTA images. While PVA is a form of partial annotation (PA) which has been adopted by a number of researches \cite{cheng2020self,ho2021deep,peng2019semi,10044712}, our proposed PVA differs from the commonly used PA methods. More specifically, PVA labels vessels continuously from the proximal end to the distal end, while the labeled regions of PA are typically randomly selected. Thus, our proposed PVA has two merits. 1) PVA balances efficiency and informativity. Compared with full annotation, PVA only requires clinicians to label vessels within restricted regions in adjacent slices, rather than all scattered target regions in each individual slice. Compared with PA, PVA keep labeled vessels continuous to preserve local topology information. 2) PVA provides flexibility for clinicians. Given that clinical diagnosis places greater emphasis on the trunks rather than the branches, PVA allows clinicians to focus their labeling efforts on vessels of particular interest. Therefore, our proposed PVA is well-suited for clinical use.

In this paper, we further propose a progressive weakly supervised learning framework for PVA. Our proposed framework, using PVA (only 24.29\% vessels labeled), achieved better performance than the competing weakly supervised methods, and comparable performance in trunk continuity with the full annotation (100\% vessels labeled) supervised baseline model. The framework works in two stages, which are local feature extraction (LFE) stage and global structure reconstruction (GSR) stage. 1) LFE stage extracts the local features of coronary artery from the limited labeled vessels in PVA, and then propagates the knowledge to unlabeled regions. 2) GSR stage leverages prediction consistency during the iterative self-training process to correct the errors, which are introduced inevitably by the label propagation process. The code of our method is available at \url{https://github.com/ZhangZ7112/PVA-CAS}.

To summarize, the contributions of our work are three-fold:

\begin{itemize}
	\item To the best of our knowledge, we proposed partial vessels annotation for coronary artery segmentation for the first time, which is in accord with clinical use. First, it balances efficiency and informativity. Second, it provides flexibility for clinicians to annotate where they pay more attention. 
	
	\item We proposed a progressive weakly supervised learning framework for partial vessels annotation-based coronary artery segmentation. It only required 24.29\% labeled vessels, but achieved comparable performance in trunk continuity with the baseline model using full annotation. Thus, it shows great potential to lower the label cost for relevant clinical and research use.
	
	\item We proposed an adaptive label propagation unit (LPU) and a learnable plug-and-play feature prototype analysis (FPA) block in our framework. LPU integrates the functions of pseudo label initialization and updating, which dynamically adjusts the updating weights according to the calculated confidence level. FPA enhances vessel continuity by leveraging the similarity between feature embeddings and the feature prototype. 
	
\end{itemize}
\section{Method}

\begin{figure*}
	\centering
	\includegraphics[width=\linewidth]{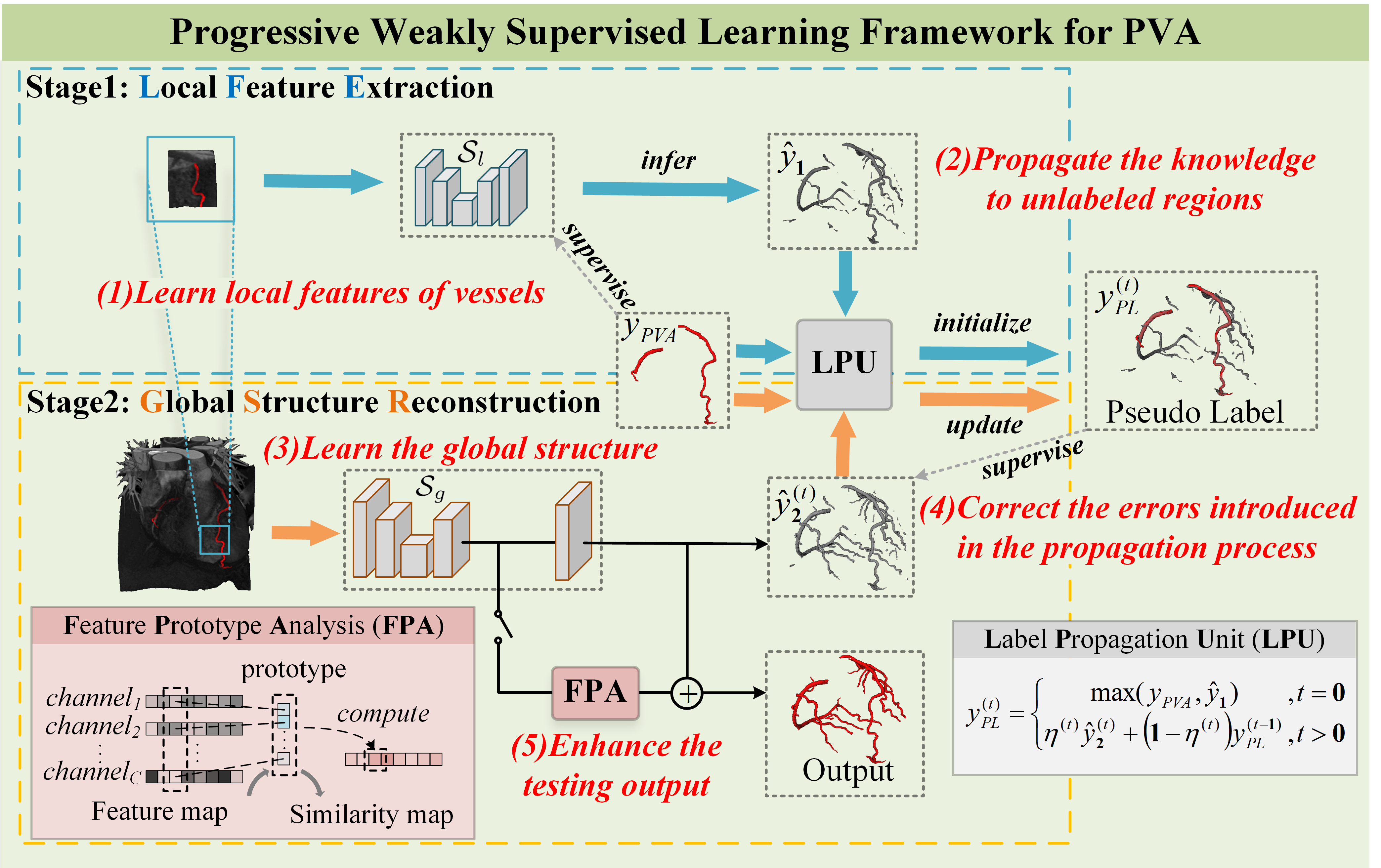}
	\caption{\textbf{Two-stage framework.} \textbf{LFE} stage: 1) $\mathcal{S}_{l}$ learns local features from the labeled vessels in PVA labels. 2) $\mathcal{S}_{l}$ propagates the knowledge to unlabeled regions and LPU initializes the pseudo labels. \textbf{GSR} stage: 3) $\mathcal{S}_{g}$ learns the global structure from the pseudo labels. 4) LPU updates the pseudo labels if a quality control test is passed. 5) FPA improves the testing output after the iterative self-training process of (3) and (4).}
	\label{fig1}
\end{figure*}

As shown in Fig.~\ref{fig1}, our proposed framework for partial vessels annotation (PVA) works in two stages. 1) The LFE stage(Sec.~\ref{subsec1}) extracts and learns vessel features from PVA locally. After the learning process, it infers on the training set to propagate the learned knowledge to unlabeled regions, outputs of which are integrated with PVA labels to initialize pseudo labels. 2) The GSR stage (Sec.~\ref{subsec2}) utilizes pseudo labels to conduct self-training, and leverages prediction consistency to improve the pseudo labels. In our proposed framework, we also designed an adaptive label propagation unit (LPU) and a learnable plug-and-play feature prototype analysis (FPA) block. LPU initialize and update the pseudo labels; FPA block learns before testing and improves the final output during testing.

\subsection{Local Feature Extraction Stage}
\label{subsec1}
In LFE stage, our hypothesis is that the small areas surrounding the labeled regions hold valid information. Based on this, a light segmentation model $\mathcal{S}_{l}$ is trained to learn vessel features locally, with small patches centering around the labeled regions as input and output. In this manner, the negative impact of inaccurate supervision information in unlabeled regions is also reduced.

\subsubsection{Pseudo Label Initialization in LPU.} After training, $\mathcal{S}_{l}$ propagates the learned knowledge of local feature to unlabeled regions. For each image of shape $H\times W\times D$, the corresponding output logit $\hat{y}_{1}\in [0,1]^{H\times W\times D}$ of $\mathcal{S}_{l}$ provides a complete estimate of the distribution of vessels, albeit with some approximation. Meanwhile, the PVA label ${y}_{PVA}\in \{0,1\}^{H\times W\times D}$ provides accurate information on the distribution of vessels, but only to a limited extent. Therefore, LPU integrate $\hat{y}_{1}$ and ${y}_{PVA}$ to initialize the pseudo label $y_{PL}$ (Equ.~\ref{equ:init_pseudo}), which will be utilized in GSR stage and updated during iterative self-training.


\begin{equation}
\label{equ:init_pseudo}
\mathop{y_{PL}^{(t=0)}(h,w,d)}\limits_{\forall (h,w,d) \in \mathbb{R}^{H\times W\times D}}= 
\begin{cases}
1, &{y}_{PVA}(h,w,d)=1,\\
\hat{y}_{1}(h,w,d), &\text{otherwise}.
\end{cases}
\end{equation}

\subsection{Global Structure Reconstruction Stage}
\label{subsec2}
The GSR stage mainly consists of three parts: 1) The segmentation model $\mathcal{S}_{g}$ to learn the global tree-like structure; 2) LPU to improve pseudo labels; 3) FPA block to improve segmentation results at testing.

Through initialization (Equ.~\ref{equ:init_pseudo}), the initial pseudo label $y_{PL}^{(t=0)}$ contains the information of both PVA labels and the knowledge of local features in $\mathcal{S}_{l}$. Therefore, at the beginning of this stage, $\mathcal{S}_{g}$ learns from $y_{PL}^{(t=0)}$ to warm up. After this, logits of $\mathcal{S}_{g}$ are utilized to update the pseudo labels during iterative self-training.

\subsubsection{Pseudo Label Updating in LPU.} The principle of this process is that more reliable logit influences more the distribution of the corresponding pseudo label. Based on this principle, first we calculate the confidence degree $\eta^{(t)}\in [0,1]$ for $\hat{y}_{2}^{(t)}$. Defined by Equ.~\ref{equ:eta}, $\eta^{(t)}$ numerically equals to the average of the logits in labeled regions. This definition makes sense since the expected logit equals to ones in vessel regions and zeros in background regions. The closer $\hat{y}_{2}^{(t)}$ gets to the expected logit, the higher $\eta^{(t)}$ (confidence degree) will be.

\begin{equation}
\label{equ:eta}
\eta^{(t)} = \frac{\sum\limits_{h}\sum\limits_{w}\sum\limits_{d}{y}_{PVA}(h,w,d) \cdot \hat{y}_{2}^{(t)}(h,w,d)}
{\sum\limits_{h}\sum\limits_{w}\sum\limits_{d} {y}_{PVA}(h,w,d)}
\end{equation}

Then, a quality control test is performed to avoid negative optimization as far as possible. If the confidence degree $\eta^{(t)}$ is higher than all elements in the set $\{\eta^{(i)} \}_{i=1}^{t-1}$, the current logit is trustworthy to pass the test to improve the pseudo label. Then, $y_{PL}^{(t)}$ is updated by the exponentially weighted moving average (EWMA) of the logits and the pseudo labels (Equ.~\ref{equ:update}). This process is similar to prediction ensemble \cite{nguyen2019self}, which hase been adopted to filter pseudo labels\cite{lee2020scribble2label}. However, different from their methods, where the factor $\eta^{(t)}$ is a fixed hyperparameter coefficient and the pseudo labels are updated each or every several epoches, $\eta^{(t)}$ in our method is adaptive and a quality control test is performed. 

\begin{equation}
\label{equ:update}
y_{PL}^{(t)}=
\begin{cases}
\eta^{(t)}\hat{y}_{2}^{(t)}+(1-\eta^{(t)})y_{PL}^{(t-1)}, &\eta^{(t)}=max\{\{\eta^{(i)}\}_{i=1}^{t} \} \\
y_{PL}^{(t-1)}, &\text{otherwise}.
\end{cases}
\end{equation}

\subsubsection{Feature Prototype Analysis Block.} Inspired by \cite{zhang2021weakly}, which generates class feature prototype $\rho _{c}$ (Equ.~\ref{equ:rho_c}) from the embeddings $z^{l}_{i}$ of labeled points in class $c$, we inherit the idea but further transform the mechanism into the proposed learnable plug-and-play block, FPA block. Experimental experience finds that the output of FPA block has good continuity, for which the FPA output are utilized to enhance the continuity of convolution output at testing.

\begin{equation}
\label{equ:rho_c}
\rho _{c} = \frac{1}{\vert \mathcal{I}_c \vert}\sum\limits_{z^{l}_{i}\in\mathcal{I}_c}z^{l}_{i}
\end{equation}

In the penultimate layer of the network, which is followed by a $1\times1\times1$ convolutional layer to output logits, we parallelly put the feature map $Z\in\mathcal{R}^{C\times H\times W\times D}$ into FPA. The output similarity map $O\in\mathcal{R}^{1\times H\times W\times D}$ is calculated by Equ.~\ref{equ:FPA}, where $Z(h,w,d)\in \mathcal{R}^{C}$ denotes the feature embeddings of voxel $(h,w,d)$, and $\rho_{\theta} \in \mathcal{R}^{C}$ the kernel parameters of FPA.

\begin{equation}
\label{equ:FPA}
O(h,w,d)=exp(-\Vert Z(h,w,d)-\rho_{\theta} \Vert^{2})
\end{equation}

The learning process of FPA block is before testing, during which the whole model except FPA gets frozen. To reduce the additional overhead, $\rho_{\theta}$ is initialized by one-time calculated $\rho _{c}$ and fine-tuned with loss $\mathcal{L}_{fpa}$ (Equ.~\ref{equ:L_fpa}), where only labeled voxels will take effect in updating the kernel.

\begin{equation}
\label{equ:L_fpa}
\mathcal{L}_{fpa}=\frac{\sum\limits_{h}\sum\limits_{w}\sum\limits_{d} {y}_{PVA}(h,w,d)\cdot log(O(h,w,d))}{\sum\limits_{h}\sum\limits_{w}\sum\limits_{d} {y}_{PVA}(h,w,d)}
\end{equation}

\section{Experiments and Results}
\subsection{Dataset and Evaluation Metrics}
Experiments are implemented on a clinical dataset, which includes 108 subjects of CCTA volumes (2:1 for training and testing). The volumes share the size of $512 \times 512 \times D$, with $D$ ranging from 261 to 608. PVA labels of the training set are annotated by clinicians, where only $24.29\%$ vessels are labeled.

The metrics used to quantify the results include both integrity and continuity assessment indicators. Integrity assessment indicators are Mean Dice Coefficient (Dice), Relevant Dice Coefficient (RDice) \cite{qi2021examinee}, Overlap (OV) \cite{kiricsli2013standardized}; continuity assessment indicators are Overlap util First Error (OF) \cite{schaap2009standardized} on the three main trunks (LAD, LCX and RCA).
\subsection{Implementation Details}
3D U-Net\cite{cciccek20163d} is set as our baseline model. Experiments were implemented using Pytorch on GeForce RTX 2080Ti. Adam optimizer was used to train the models with an initial learning rate of $10^{-4}$. The patch sizes were set as $128 \times 128 \times 128$ and $512 \times 512 \times 256$ respectively for $\mathcal{S}_{l}$ and $\mathcal{S}_{g}$. When testing, sliding windows were used with a half-window width step to cover the entire volume.

\subsection{Comparative Test}
To verify the effectiveness of our proposed method, it is compared  with both classic segmentation models (3D U-Net \cite{cciccek20163d}, HRNet \cite{sun2019deep}, Transunet \cite{chen2021transunet}) and partial annotation-related weakly supervised frameworks (EWPA \cite{peng2019semi}, DMPLS \cite{luo2022scribble}).

\begin{table*}
	\centering
	\caption{Quantative results of different methods under partial vessels annotation (PVA, 24.29\% vessels labeled) or full annotation (FA, 100\% vessels labeled).}
	
	\resizebox{1.0\textwidth}{!}{
	\begin{tabular}{c|l|c|c|c|c|c|c}
		\hline
		\multirow{2}{*}{Label}&\multicolumn{1}{c|}{\multirow{2}{*}{Method}} &\multirow{2}{*}{Dice(\%)$\uparrow$} &\multirow{2}{*}{RDice(\%)$\uparrow$} 
		&\multirow{2}{*}{OV(\%)$\uparrow$} &   	\multicolumn{3}{c}{OF$\uparrow$}            \\
		\cline{6-8}  		            				          				
		& & & & &    					LAD 		&LCX      		&RCA          \\
		\hline
		\multirow{6}{*}{PVA}&3D U-Net \cite{cciccek20163d}			& 60.60$_{\pm7.09}$              &69.45$_{\pm7.82}$
		&62.24$_{\pm6.43}$			&0.647$_{\pm0.335}$&0.752$_{\pm0.266}$&0.747$_{\pm0.360}$ \\
		\cline{2-8} 
		&HRNet \cite{sun2019deep} 				&48.72$_{\pm7.16}$		&52.31$_{\pm7.96}$
		&37.81$_{\pm6.61}$			&0.490$_{\pm0.297}$&0.672$_{\pm0.301}$	&0.717$_{\pm0.356}$\\
		\cline{2-8} 
		&Transunet \cite{chen2021transunet}		&63.08$_{\pm6.42}$		&71.97$_{\pm7.38}$
		&61.21$_{\pm6.40}$		&0.669$_{\pm0.274}$&0.762$_{\pm0.243}$	&0.728$_{\pm0.362}$\\
		\cline{2-8} 
		&EWPA \cite{peng2019semi}				&55.41$_{\pm6.15	}$	&61.54$_{\pm6.83}$		
		&60.48$_{\pm5.17	}$		&0.659$_{\pm0.334}$&0.759$_{\pm0.286}$	&0.749$_{\pm0.364}$\\
		\cline{2-8} 
		&DMPLS \cite{luo2022scribble}			&59.12$_{\pm7.69}$	&65.81$_{\pm8.15}$
		&59.99$_{\pm5.80}$			&0.711$_{\pm0.292}$&0.775$_{\pm0.284}$	&0.711$_{\pm0.358}$\\
		\cline{2-8} 
		&\textbf{Ours}							&\textbf{71.45$_{\pm\mathbf{6.07}}$} 	&\textbf{83.14$_{\pm\mathbf{6.72}}$}	
		&\textbf{75.40$_{\pm\mathbf{6.15}}$}		&\textbf{0.895$_{\pm\mathbf{0.226}}$}&\textbf{0.915$_{\pm\mathbf{0.190}}$}	&\textbf{0.879$_{\pm\mathbf{0.274}}$}\\
		\hline
		FA&3D U-Net&83.14$_{\pm3.52	}$	&90.91$_{\pm4.18}$
		&89.00$_{\pm4.75}$			&0.913$_{\pm0.231}$&0.843$_{\pm0.301}$	&0.873$_{\pm0.265}$\\
		\hline
\end{tabular}}\label{tab1}
\end{table*}

\begin{figure*}
	\centering
	\includegraphics[width=\linewidth]{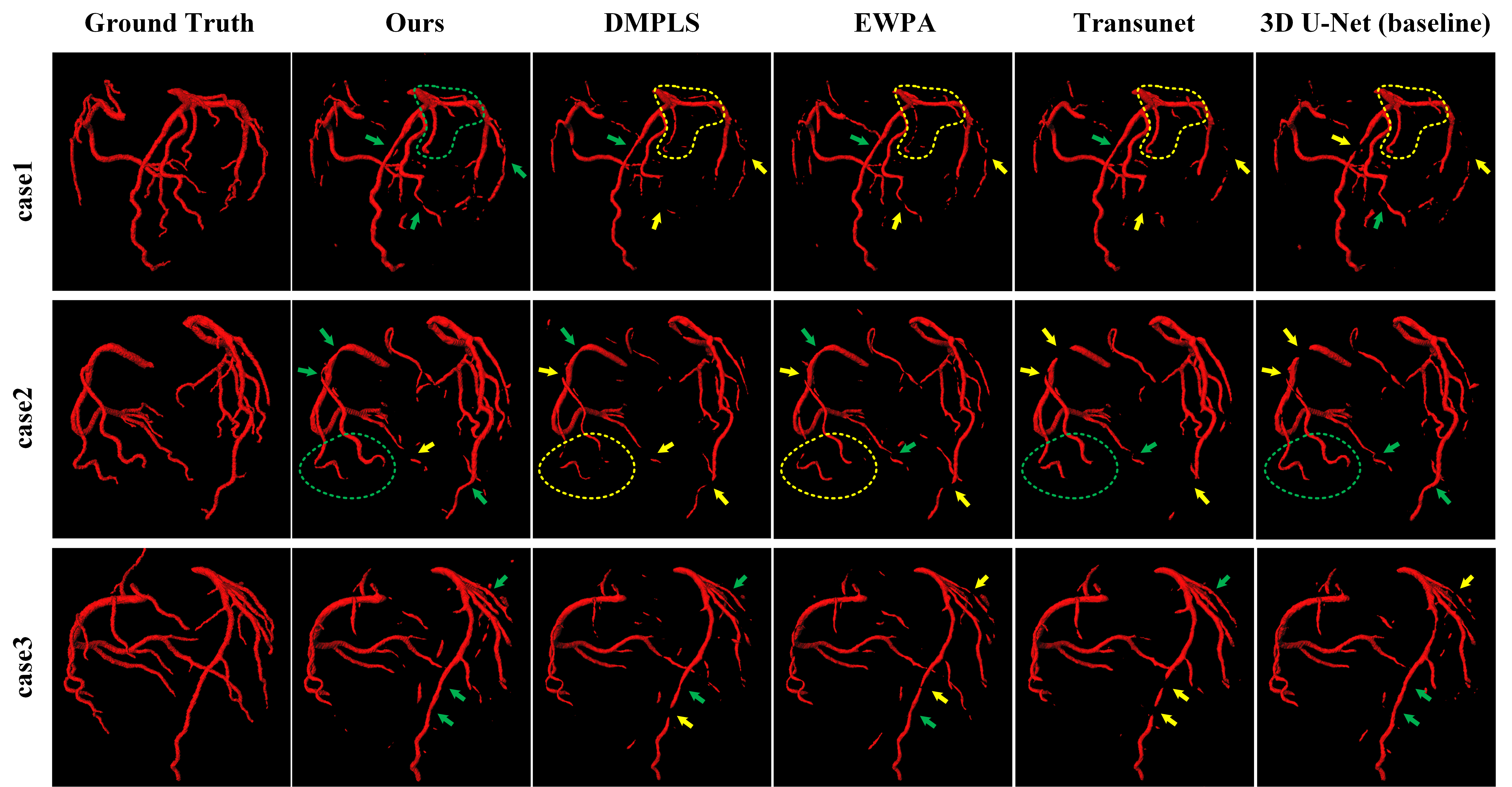}
	\caption{Visual comparison of the segmentation results under PVA. Green symbols (arrows and dotted frames) indicate higher-quality regions than yellow symbols.}
	\label{fig3}
\end{figure*}

The quantative results of different methods are summarized in Tab.~\ref{tab1}, which shows that our proposed method outperforms the competing methods under PVA. The competing frameworks (EWPA and DMPLS) had achieved the best results in their respective tasks under partial annotation, but our proposed method achieved better results for PVA-based coronary artery segmentation. It is worth mentioning that the performance in trunk continuity (measured by the indicator OF) of our proposed method using PVA (24.29\% vessels labeled) is comparable to that of the baseline model using full annotation (100\% vessels labeled).

The qualitative visual results verify that our proposed method outperforms the competing methods under PVA. Three cases are shown in Fig.~\ref{fig3}. All the cases show that the segmentation results of our method have good overall topology integrity, especially on trunk continuity.

\subsection{Ablation Study}

Ablation experiments were conducted to verify the importance of the components in our proposed framework (summarized in Tab.~\ref{tab2}). The performance improvement verifies the effectiveness of pseudo label initialization (PLI) and updating (PLU) mechanisms in the label propagation unit (LPU). PLI integrates the information of PVA labels with the propagated knowledge, and PLU improves the pseudo labels during self-training. With the help of FPA block, the segmentation results gain further improvement, especially on the continuity of trunks.

\begin{table*}
	\centering
	\caption{Quantative results of ablation analysis of different components.}
	
	\resizebox{1.0\textwidth}{!}{
		\begin{tabular}{c|c|c|c|c||c|c|c|c|c|c}
			\hline
			\multirow{2}{*}{$\mathcal{S}_{l}$}&\multicolumn{2}{c|}{LPU} & \multirow{2}{*}{$\mathcal{S}_{g}$}& \multirow{2}{*}{FPA} &\multirow{2}{*}{Dice(\%)$\uparrow$} &\multirow{2}{*}{RDice(\%)$\uparrow$} &\multirow{2}{*}{OV(\%)$\uparrow$} &   	\multicolumn{3}{c}{OF$\uparrow$}            \\
			\cline{2-3} \cline{9-11}	            				          				
			&PLI & PLU &          &  &   & & &   					LAD 		&LCX      		&RCA          \\
			\hline
			
			\checkmark&	&	&	&			
			& $60.60_{\pm7.09}$             &$69.45_{\pm7.82}$
			&$62.24_{\pm6.43}$			&$0.647_{\pm0.335}$&$0.752_{\pm0.266}$&$0.747_{\pm0.360}$ \\
			\hline
			\checkmark&\checkmark	&	&\checkmark	&  				
			&$64.23_{\pm6.44}$		&$73.81_{\pm6.89}$
			&$66.19_{\pm5.63}$		&$0.751_{\pm0.328}$&$0.813_{\pm0.231}$	&$0.784_{\pm0.349}$\\
			\hline
			\checkmark&\checkmark	&\checkmark	&\checkmark	& 		
			&$71.43_{\pm7.20}$		&$81.70_{\pm6.92}$
			&$72.13_{\pm5.94}$		&$0.873_{\pm0.227}$&$0.860_{\pm0.223}$	&$0.808_{\pm0.334}$\\
			\hline
			\checkmark&\checkmark	&\checkmark	&\checkmark	&\checkmark 			
			&\textbf{71.45$_{\pm\mathbf{6.07}}$}	&\textbf{83.14$_{\pm\mathbf{6.72}}$}
			&\textbf{75.40$_{\pm\mathbf{6.15}}$}		&\textbf{0.895$_{\pm\mathbf{0.226}}$}&\textbf{0.915$_{\pm\mathbf{0.190}}$}&\textbf{0.879$_{\pm\mathbf{0.274}}$}\\
			\hline
			
		\end{tabular}
	}
	
	\label{tab2}
\end{table*}

\section{Conclusion}
In this paper, we proposed partial vessels annotation (PVA) for coronary artery segmentation on CCTA images. The proposed PVA is convenient for clinical use for the two merits, providing flexibility as well as balancing efficiency and informativity. Under PVA, we proposed a progressive weakly supervised learning framework, which outperforms the competing methods and shows comparable performance in trunk continuity with the full annotation supervised baseline model. In our framework, we also designed an adaptive label propagation unit (LPU) and a learnable plug-and-play feature prototype analysis(FPA) block. LPU integrates the functions of pseudo label initialization and updating, and FPA improves vessel continuity by leveraging the similarity between feature embeddings and the feature prototype. To conclude, our proposed framework under PVA shows great potential for accurate coronary artery segmentation while requiring significantly less annotation effort.

%
%
%
\bibliographystyle{splncs04}
\bibliography{selfbib}
\end{document}